\PassOptionsToPackage{table}{xcolor}
\documentclass[]{gtech}
\usepackage{amssymb}
\usepackage{multirow}
\usepackage{bigdelim}
\usepackage{todonotes}
\usepackage{longtable}
\usepackage{tabularray}
\usepackage{wrapfig}
\usepackage[most]{tcolorbox} 
\usepackage[table]{xcolor}
\usepackage{url}
\usepackage{xspace}

\renewcommand{\title}[1]{\newcommand{\titlelist}{{\huge\fontfamily{optimistic}\selectfont #1}}}

\newcommand{\method}{M2-Reasoning\xspace}

\definecolor{prompt}{HTML}{5f84e4}
\definecolor{img}{HTML}{820100}

\usepackage{arydshln}

\definecolor{CQColor}{rgb}{0.0,0.0,1.0} 

\definecolor{TSColor}{rgb}{0.5,0.0,0.8} 

\definecolor{CQRColor}{rgb}{1.0,0.0,1.0} 

\definecolor{lightblue}{RGB}{227, 244, 254}

\usepackage{colortbl}
\usepackage{amssymb}
\usepackage{pifont}
\usepackage{booktabs,multirow}
\usepackage{makecell}
\usepackage{tabulary}
\usepackage{fontawesome5}
\usepackage{bbding}
\usepackage{multicol}
\usepackage{enumitem}
\usepackage{fancyvrb}
\usepackage{tikz}
\usepackage{lipsum}
\usepackage{subcaption}

\newlength\savewidth

\title{\textcolor[HTML]{0369ff}{\method}: Empowering MLLMs with Uni{f}{i}ed General and Spatial Reasoning}
\author[*]{Inclusion AI, Ant Group}

\contribution[*]{See Contributions section (Sec.~\ref{sec:contributors}) for full author list.}

\abstract{\fontsize{11pt}{12pt} \textit{Recent advancements in Multimodal Large Language Models (MLLMs), particularly through Reinforcement Learning with Verifiable Rewards (RLVR), have significantly enhanced their reasoning abilities. However, a critical gap persists: these models struggle with dynamic spatial interactions, a capability essential for real-world applications. To bridge this gap, we introduce \textbf{\method-7B}, a model designed to excel in both general and spatial reasoning. Our approach integrates two key innovations: (1) a novel data pipeline that generates 294.2K high-quality data samples (168K for cold-start fine-tuning and 126.2K for RLVR), which feature logically coherent reasoning trajectories and have undergone comprehensive assessment; and (2) a dynamic multi-task training strategy with step-wise optimization to mitigate conflicts between data, and task-specific rewards for delivering tailored incentive signals. This combination of curated data and advanced training allows \method-7B to set a new state-of-the-art (SOTA) across 8 benchmarks, showcasing superior performance in both general and spatial reasoning domains.
}}

\date{July 10, 2025\vspace{-1mm}}
\gtechdata[Model]{\url{https://huggingface.co/inclusionAI/M2-Reasoning}}
\gtechdata[Code]{\url{https://github.com/inclusionAI/M2-Reasoning}}

\begin{document}
\maketitle

\begin{figure}[b!]
    \centering
    \includegraphics[width=\textwidth]{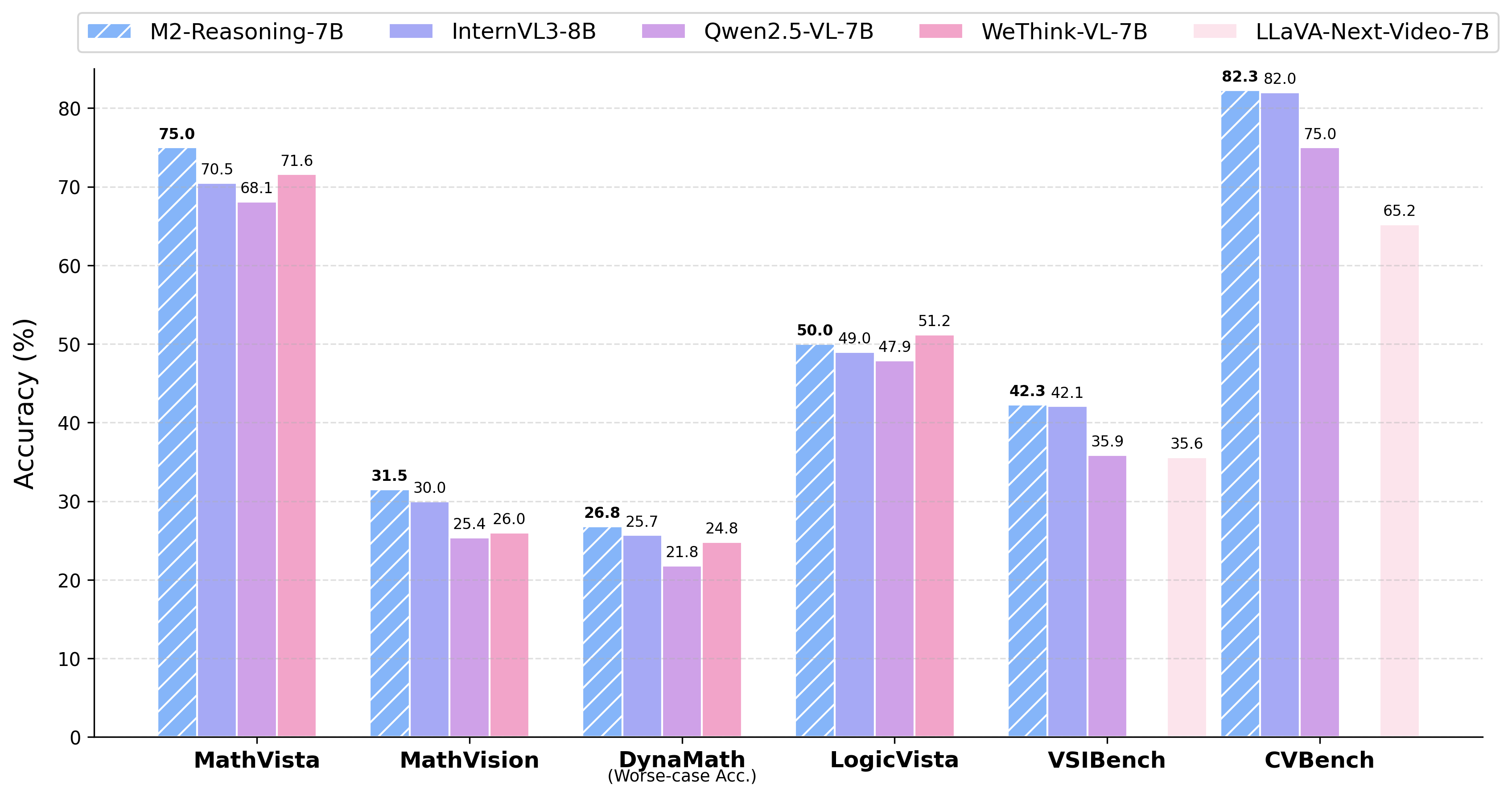}
    \vspace{-15pt}
    \caption{Benchmark performance of \method-7B.}
    \label{fig:bottom_fig}
\end{figure}

\section{Introduction}\label{sec:introduction}

The recent advancements in Large Language Models (LLMs) have redefined the landscape of artificial intelligence, with models demonstrating remarkable capabilities in language understanding, generation, and reasoning~\citep{llama3, emu3, qwen3, seed1.5, deepseek-v3, deepseek-r1, ling, cai2024internlm2, team2023gemini}. This progress has directly catalyzed the emergence of Multimodal Large Language Models (MLLMs), which integrate these sophisticated language faculties with visual perception~\citep{li2024llavaonevisioneasyvisualtask, gpt4o, qwen2.5-vl, zhu2025internvl3, li2024llavanextinterleavetacklingmultiimagevideo}. Consequently, MLLMs can now adeptly handle a diverse spectrum of complex vision-language tasks, such as visual question answering, detailed image captioning, object grounding, and nuanced visual reasoning~\citep{jaech2024openaio1, Mingomni2025, seed1.5-vl, kimivl}.

Recent breakthroughs in leading LLMs, such as OpenAI's o1/o3~\citep{jaech2024openaio1} and Deepseek-R1~\citep{deepseek-r1} have shown that a two-stage training process involving Supervised Fine-Tuning (SFT) and Reinforcement Learning with Verifiable Rewards (RLVR) is highly effective for unlocking advanced reasoning abilities. Inspired by this success, the same paradigm is increasingly being adapted for MLLMs to bolster their cognitive capabilities~\citep{vlaa-thinker, wethink, Observe-R1, lmm-r1, deng2025openvlthinker}. However, a critical gap remains: these models largely fail to reason about the dynamic interplay of space and motion, a capability essential for meaningful real-world applications.

To address this critical gap, we introduce \textbf{\method-7B}, a Multimodal Large Language Model (MLLM) engineered to excel in two fundamental domains: \textbf{general reasoning}, for solving complex abstract problems, and \textbf{spatial reasoning}, for comprehending motion, orientation, and physical interactions in dynamic environments~\citep{ray2025satdynamicspatialaptitude}.

The development of \method is underpinned by a synergistic combination of a novel data curation strategy and an advanced training framework. 

First, to overcome the scarcity of high-quality data, we established a multi-stage pipeline for data synthesis and curation. This process includes automated question-answer generation and the synthesis of detailed reasoning trajectories. Critically, all generated data undergoes a holistic evaluation that assesses not only correctness but also its quality, diffucilty, and diversity. This comprehensive assessment framework is the key to enabling effective curriculum learning during the RLVR stage, as it allows us to strategically sequence the training data from simple to complex. This pipeline yielded 168K instances for cold-start, and a further 126.2K for RLVR, comprising 100K for general reasoning and 26.2K (18.7K image-based and 7.5K video-based) for spatial reasoning.

Second, to effectively train our model on this diverse dataset, we introduce a sophisticated training strategy that combines multi-task optimization with domain specialization. Our approach employs step-wise dynamic optimization to mitigate conflicts arising from data heterogeneity, and avoid suboptimal, one-size-fits-all configurations. Furthermore, we incorporate a task-specific reward formulation, which designs distinct reward functions tailored to the unique requirements of each task category. This is particularly crucial for spatial reasoning since such tailored incentive signals are essential for effectively bootstrapping the MLLM's nascent perceptual abilities.

In summary, the contributions of this work are three-fold:

\begin{enumerate}[label=$\blacklozenge$, wide=0pt, leftmargin=*]
\item \textbf{A High-quality Data Construction Pipeline:} We design and implement a multi-stage data synthesis and curation pipeline that generates vast amounts of reasoning data. (See Section~\ref{sec:data})

\item \textbf{A Dynamic Multi-Task Training Strategy:} We propose a sophisticated training strategy that effectively handles data heterogeneity. It features step-wise dynamic optimization to mitigate conflicts between different data sources and a task-specific reward formulation to provide tailored incentive signals. (See Section~\ref{sec:approach})

\item \textbf{Unified General and Spatial Reasoning Model:} We propose \method-7B, an MLLM uniquely engineered for both abstract and spatial reasoning. Extensive evaluations on 8 distinct benchmarks demonstrate that, by leveraging our custom data and training pipelines, \method-7B establishes new state-of-the-art (SOTA) results across both general and spatial reasoning domains. (See Section~\ref{sec:experiments})

\end{enumerate}
\section{Data Construction}\label{sec:data}

To comprehensively activate and enhance the model’s reasoning capabilities across the tasks, we propose a \textbf{high-quality data construction pipeline} that incorporates general reasoning and spatial reasoning.

Specifically, for the general reasoning tasks (Section~\ref{sec:general_reasoning_data}), we construct both cold-start and RLVR data: the former aims to help the model understand structured reasoning patterns, while the latter focuses on optimizing reasoning trajectories. For the spatial reasoning task (Section~\ref{sec:spatial_reasoning_data}), we utilize image-text and video-based for RLVR to better capture dynamic spatial transformations and temporal dependencies.

In the following sections, we provide detailed descriptions of the data construction methodologies tailored to each task type, highlighting how they contribute to the development of multi-faceted reasoning abilities in the MLLM. Finally, an overview of the overall data configuration used in our training is provided in Section~\ref{sec:overview_of_data}.

\subsection{General Reasoning Data}
\label{sec:general_reasoning_data}

To enhance the model's general reasoning capabilities, we optimize the data strategy during the cold-start stage and RLVR curriculum. First, we implement a high-quality, multimodal chain-of-thought (CoT) synthesis pipeline that generates and filters reasoning chains based on answer accuracy and fine-grained reasoning quality metrics, ensuring structured and effective training. Second, we introduce a prompt difficulty scoring method for RLVR, enabling the model to learn progressively from tasks of increasing difficulty.

\subsubsection{High-Quality Multimodal Chain-of-Thought Synthesis Pipeline}

In our preliminary experiments, we collect the open-source image-text cold-start dataset R1-OneVision~\citep{r1-onevision} and text-only data from Ling-Lite~\citep{ling} to activate the model’s general reasoning capabilities. These datasets provide a basic foundation for developing the model’s thinking ability. However, the quality and formatting of the reasoning chains contained in these datasets are not consistently controlled, which may lead to suboptimal performance when training models for complex reasoning tasks.

Previous studies~\citep{limo, mimovl} have demonstrated that high-quality thought processes significantly contribute to effective model learning. Motivated by this, we design a comprehensive pipeline for thought chain generation and quality screening. This pipeline enables us to construct well-structured, high-quality reasoning data, thereby enhancing the model's ability to perform multi-step, complex reasoning.

We begin by curating a collection of publicly available datasets~\citep{mm-eureka, geometry3k, k12-free-form, vlaa-thinker, thinklite-vl}, which span a diverse range of reasoning tasks — from geometric problem-solving to logical inference in visual environments. To generate high-quality multimodal CoT data, we utilize an open-source vision-language reasoning model, WeThink-VL-7B~\citep{wethink}, to synthesize reasoning chains for each input.

To ensure diversity in the generated reasoning chains, we sample multiple responses per instance using a high-temperature generation strategy. This approach encourages varied thought processes and enhances both the breadth and richness of the resulting CoT dataset. Next, we assess the quality of the synthesized CoT data along two key dimensions:

\begin{enumerate}[label=$\blacklozenge$, wide=0pt, leftmargin=*]
    \item \textbf{\textit{Answer Accuracy Assessment.}} We first evaluate the final answer of each reasoning chain to determine its correctness. Any reasoning chain that leads to an incorrect final answer is considered flawed in its overall thought process and is therefore discarded. To assess answer accuracy, we utilize Qwen2.5-7B-Instruct~\citep{qwen2.5}, which serves as an automated evaluator for response correctness serves as an automated evaluator for response correctness.
    \item \textbf{\textit{Reasoning Quality Assessment.}} Inspired by \citep{limo}, beyond basic answer accuracy, we further establish a quality scoring framework to identify and retain high-quality reasoning chains that are more effective at eliciting deep thinking from the model. This assessments focuses on three key dimensions:
    \begin{enumerate}[label=$\lozenge$, wide=0pt, leftmargin=*]
        \item \textit{Structural Correctness of Reasoning:}
        Assess whether the reasoning process is well-structured and clearly organized. Does it elaborate on complex steps in detail while keeping simpler ones concise, thereby demonstrating an adaptive adjustment of reasoning granularity?
        \item \textit{Cognitive Load Management:}  
        Assess whether the reasoning effectively guides the reader through complex logic by progressively introducing concepts, clearly articulating key insights, and bridging cognitive gaps.
        \item \textit{Verification Richness:}
        Assess whether the reasoning process frequently validates intermediate results, examines the validity of underlying assumptions, and ensures logical consistency throughout the inference chain.
    \end{enumerate}
    We employ Qwen2.5-7B-Instruct~\citep{qwen2.5} model, in conjunction with carefully designed prompts based on the aforementioned assessment principles, to assess the quality of the generated reasoning chains. 
\end{enumerate}

After this filtering process, we select a final set of 168K high-quality reasoning chains as our cold-start training data. Experimental results demonstrate that the data generated through this pipeline significantly enhances the model’s overall reasoning capabilities. The prompts used for scoring are provided in Appendix~\ref{app:prompts}.

\subsubsection{Difficulty Scoring for Data in Reinforcement Learning}
For the RLVR data, we quantify the difficulty of each prompt by the accuracy of model responses. Specifically, we generate ten reasoning-augmented responses per prompt using Qwen2.5-VL-7B~\citep{qwen2.5-vl} and compute the \textit{response accuracy} as the proportion of correct answers among these responses. Prompts yielding 0\% or 100\% accuracy are excluded from the final dataset, as these extremes represent either overly challenging or trivial cases that fail to provide meaningful gradient signals during RLVR training. For the remaining prompts, we define their \textit{difficulty} as $1 - \text{accuracy}$, where higher values correspond to greater task difficulty.

\subsection{Spatial Reasoning Data}
\label{sec:spatial_reasoning_data}

To improve the model’s perceptual understanding of real-world images and videos, we design a spatial reasoning dataset based on controlled spatial simulations. Notably, the video simulation data is specifically crafted to capture dynamic, video-dependent information such as room size and object appearance order.

We introduce a comprehensive \textbf{spatial data synthesis pipeline}, which ensures the creation of high-quality, semantically meaningful data tailored for training and evaluating spatial reasoning capabilities. Furthermore, we design a \textbf{spatial data augmentation pipeline} that enhances the diversity of questions, options, and instructions through secondary augmentation.

\subsubsection{Spatial Data Synthesis Pipeline}

\textbf{Automated Data Annotation.}
Our annotation process first performs preprocessing steps on real-world images, including depth estimation and instance segmentation, to obtain fundamental modalities such as depth maps, segmentation masks, surface normal maps, and camera parameters. Based on these, the module further derives 3D point clouds, 3D bounding boxes, and object labels, which are key informations for building spatial-related QA pairs. For simulated data, the annotation process prioritizes the use of built-in annotations, including point cloud-based instance segmentation, 3D bounding boxes, and object classification. These built-in labels in simulated data provide more accurate ground truth compared to real-world data.

\textbf{Question-Answer Generation.}
To endow the model with more comprehensive spatial understanding and reasoning abilities, we designed 10 categories of tasks. The image-based tasks include: \texttt{ObjectCount}, \texttt{SpatialRelation}, \texttt{RelativeSize}, \texttt{AbsoluteSize}, \texttt{AbsoluteDistance}, \texttt{RelativeDistance}, and \texttt{RelativeDepth}. The video-based tasks focus on dynamic spatial-temporal understanding and include: \texttt{RoomSize}, \texttt{AppearanceOrder}, and \texttt{RelativeDirection}.

For each question type, we design a corresponding QA generation algorithm that leverages the pre-obtained annotations to produce semantically relevant questions and answers. We also implement question-specific filtering mechanisms to eliminate ambiguous or invalid object-relation pairs, ensuring clarity and well-defined reasoning tasks.

Take the image-based \texttt{RelativeDistance} task as an example, we applied three filtering criteria to ensure unambiguous spatial comparisons: 1) Objects with multiple instances were removed to avoid localization ambiguity; 2) Object with planar surfaces such as \textit{Floor} and \textit{Ceiling} were excluded due to potential perceptual ambiguity; 3) Among the remaining candidates, objects with similar distances to the reference object were filtered out, and only triplets $(A, B, C)$ with clearly distinct distances were selected to construct the question: \textit{"Which is closer to $A$, $B$ or $C$?"}.

As another example, consider the video-based task \texttt{AppearanceOrder}. We first record object names and detection results frame-by-frame. Then, the following filters were applied: 1) Objects with excessively small pixel areas were discarded to reduce perceptual ambiguity; 2) Objects appearing simultaneously in the same frame or across adjacent frames were filtered out to prevent choice ambiguity. Finally, 3–4 objects were sampled from the remaining set, and temporal deduplication was performed to generate well-structured questions and answer options based on their appearance order.

\textbf{Data Quality Validation.} We adopt a hybrid quality control strategy combining model-based assessments with human verification. The model-based evaluation, performed by the Qwen2.5-VL-32B-Instruct~\citep{qwen2.5-vl} MLLM, encompassed two distinct types of assessments:
\begin{enumerate}[label=$\blacklozenge$, wide=0pt, leftmargin=*]
    \item \textbf{\textit{Difficulty Assessment.}} We fed the visual input, question, and answer into the MLLM. The model performed multiple reasoning processes, and the accuracy across responses was computed to derive a difficulty score, defined as $1-\text{accuracy}$.
    \item \textbf{\textit{Comprehensive Assessment.}} The visual input, question, and answer were jointly input into the MLLM, which independently evaluated three aspects — \textit{the correctness of the answer}, \textit{the completeness of the objects involved in the question}, and \textit{the discriminability of the answer choices} — and generated corresponding labels for each.
\end{enumerate}

After obtaining all the above labels, we conducted manual verification to ensure their accuracy and performed preliminary data filtering based on these labels. Additionally, we carried out a second round of human evaluation on a small random sample to further ensure data quality.

\subsubsection{Spatial Data Augmentation Pipeline}

Each synthesized spatial data consists of \texttt{VisualInput}, \texttt{Question}, \texttt{TargetAnswer}, and \texttt{CandidateOptions}. We perform the following key data augmentation strategies:
\begin{enumerate}[label=$\blacklozenge$, wide=0pt, leftmargin=*]
    \item \textbf{Question-Type Augmentation.} To balance the distribution of question types, we transform original multiple-choice questions into fill-in-the-blank and true/false formats based on the \texttt{TargetAnswer} and \texttt{CandidateOptions}. For instance, a portion of \texttt{ObjectCount} questions are converted into numerical fill-in-the-blank tasks, while some \texttt{SpatialRelation} questions are reformulated as true/false questions. In the latter case, a question originally asking for \textit{"left"} or \textit{"right"} becomes a binary judgment such as: \textit{"Is object $A$ to the left of object $B$?"}
    \item \textbf{Instruction Augmentation.} To preserve instruction-following performance, we augment the \texttt{Question} by enriching prompts while ensuring consistency with the \texttt{TargetAnswer}. Specifically, we embed formatting requirements—such as units, response types, or structured formats—directly into the questions. This enables the accuracy-based reward to also encourage proper adherence to instructions. For example, a question like \textit{"What is the distance between $A$ and $B$?"} may be expanded to include \textit{"measured in centimeters"} or \textit{"measured in meters"}, each paired with the corresponding correct answer.
    \item \textbf{Distribution Augmentation.} To address the bias introduced by imbalanced \texttt{TargetAnswer} distributions in the original data, we perform augmentation on the \texttt{CandidateOptions}. Specifically, we introduce a random offset to the option letters in both \texttt{TargetAnswer} and \texttt{CandidateOptions}, which helps reduce the model's reliance on specific option positions and encourages more robust and generalizable learning.
\end{enumerate}

\subsection{Overview of Data Configuration}
\label{sec:overview_of_data}

\begin{figure}[htbp]
    \centering
    \begin{subfigure}[b]{0.48\textwidth}
    \captionsetup{justification=centering}
        \includegraphics[width=\textwidth]{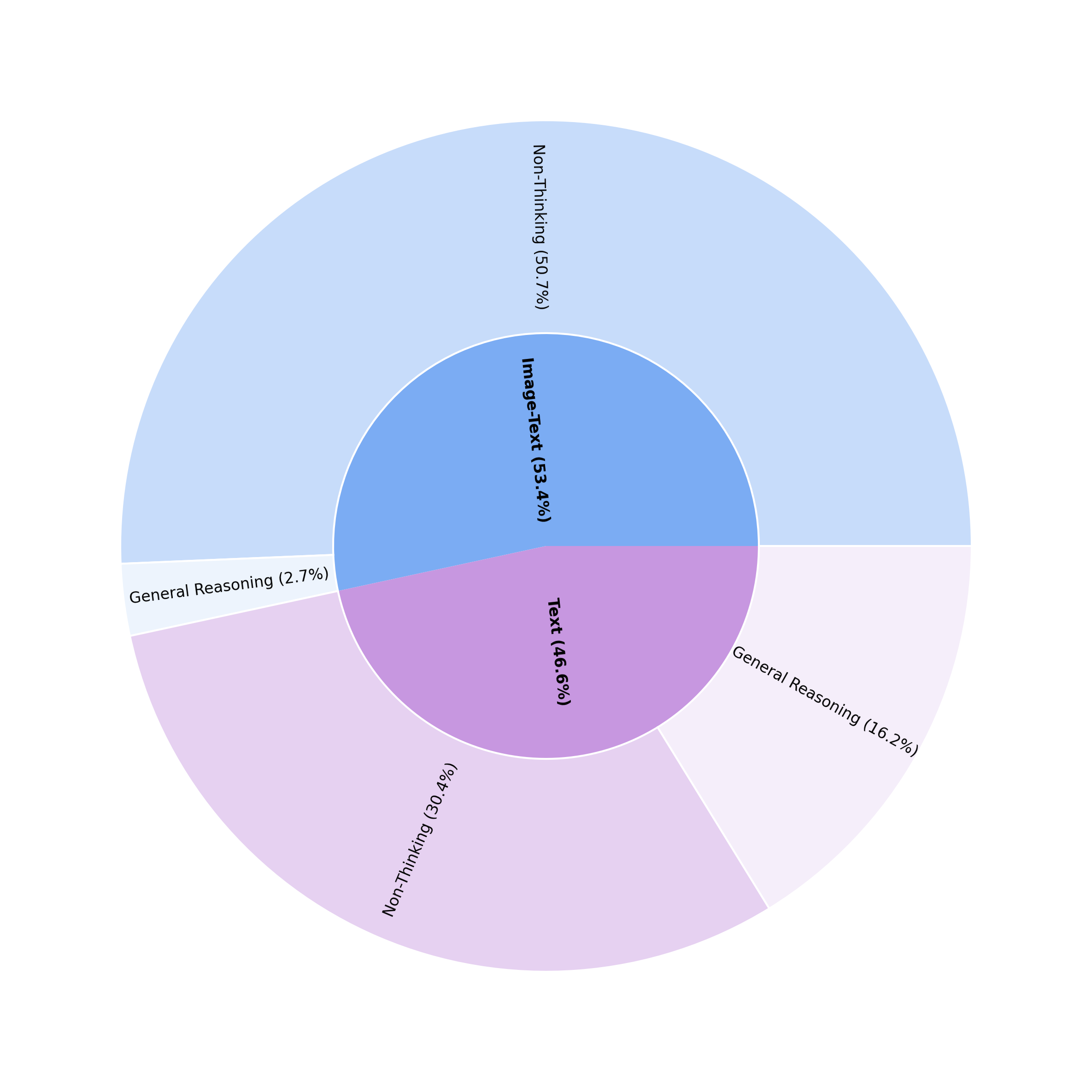}
        \caption{\small{Cold-Start Data\\Image-Text: 3.3M\\Text: 2.9M}}
        \label{fig:subfig_a}
    \end{subfigure}
    \hfill
    \begin{subfigure}[b]{0.48\textwidth}
    \captionsetup{justification=centering}
        \includegraphics[width=\textwidth]{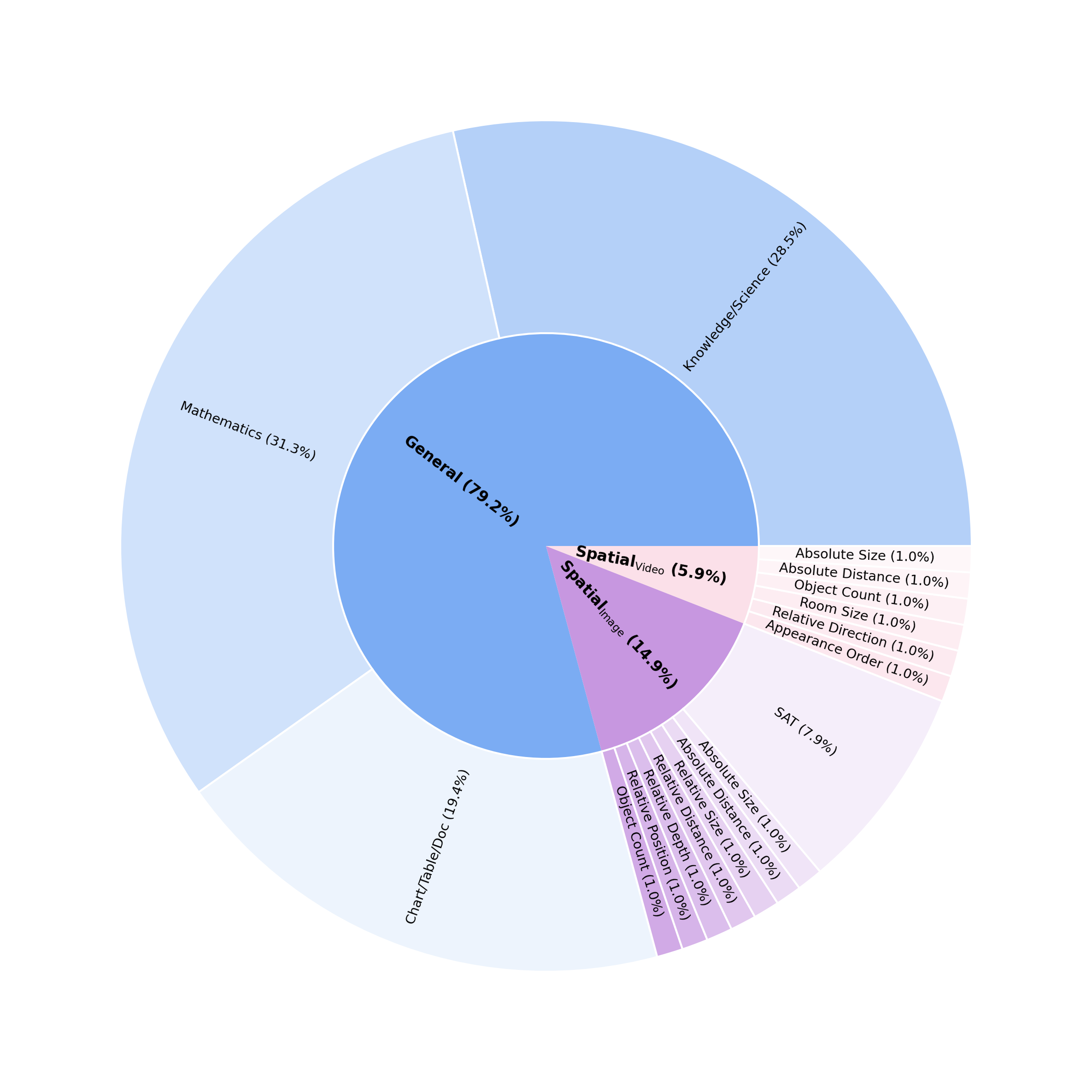}
        \caption{\small{RLVR Data\\Genral: 100K\\Spatial\textsubscript{Image}: 18.8K\\Spatial\textsubscript{Video}: 7.5K}}
        \label{fig:subfig_b}
    \end{subfigure}
    \caption{Overview of the data configurations during cold-start and RLVR.}
    \label{fig:data_config}
\end{figure}

The overview of data configurations for \method are depicted in Figure~\ref{fig:data_config}. The cold-start stage (Figure~\ref{fig:data_config}(a)) employs a large-scale dataset of 3.3 million image-text data and 2.9 million text-only samples. Within this stage, the "General Reasoning" data is sourced from Section~\ref{sec:general_reasoning_data} for the image-text modality and from Ling-Lite~\citep{ling} for the text-only modality. Additionally, to mitigate catastrophic forgetting and preserve the model's foundational capabilities, we incorporate a substantial amount of "Non-Thinking" data from the Instruction-Tuning-Stage-3 of M2-Omni~\citep{guo2025m2}. The subsequent RLVR stage (Figure~\ref{fig:data_config}(b)) utilizes a smaller, curated dataset to hone the model's advanced reasoning abilities. This dataset is heavily focused on complex tasks such as mathematics, science, and fine-grained spatial reasoning, spanning both image and video modalities.

\section{Approach}\label{sec:approach}

\begin{figure}[htbp]
    \centering
    \includegraphics[width=\textwidth]{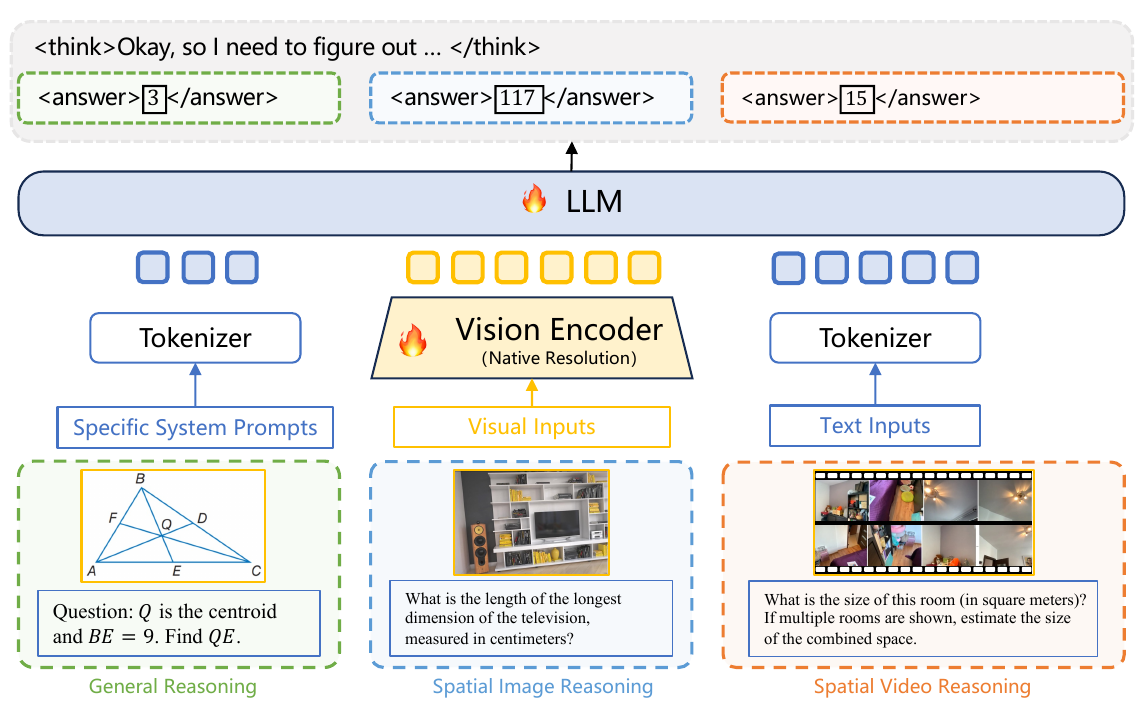}
    \vspace{-15pt}
    \caption{\textbf{The \method's model architecture} is built upon the Qwen2.5-7B language model, incorporating a native-resolution vision encoder. Notably, the figure omits the MLP projector typically used to connect the vision encoder and the language model.}
    \label{fig:model_architecture}
\end{figure}

As depicted in Figure~\ref{fig:model_architecture}, \method is a unified model capable of supporting general and spatial reasoning tasks. The development of \method comprises two core components: (1) a dynamic reasoning incentive framework that harmonizes multi-task optimization, and (2) task-specific reward formulation that preserve domain specialization. The following sections will explore how \method achieves the superior ability across all tasks.

\subsection{Dynamic Reasoning Incentivizing}
\subsubsection{Two-Stage Training Strategy}
\label{sec:two-stage-training-strategy}

\method initializes by adapting the M2-Omni~\citep{guo2025m2} framework. We leverage its established training process and data, but crucially replace the Llama-3~\citep{llama3} LLM with Qwen2.5-7B-Instruct~\citep{qwen2.5} for its enhanced language modeling capabilities.

To advance multimodal reasoning performance, we propose a two-stage training framework.

\textbf{Stage 1: Cold-start.} In this stage, supervised fine-tuning (SFT) is conducted on cold-start data to activate the model’s latent reasoning capabilities and standardize its output format. This step helps mitigate instability during the early phases of RLVR training. Our preliminary experiments align with findings from previous studies~\citep{vision-r1, r1-onevision, vlaa-thinker}, which demonstrate that initializing the training process with high-quality and semantically coherent trajectory data can enhance the upper bound of the model's reasoning performance. Upon completing this stage, we obtained the cold-start initialized model, denoted as \method-CI.

\textbf{Stage 2: Dynamic Multi-task RLVR.} In the second stage, Reinforcement Learning with Verifiable Rewards (RLVR) is applied to data with verifiable answers, encouraging the model to adopt the correct reasoning process and improve its generalization across a diverse range of multimodal tasks. We adopt GRPO for our reasoning tasks due to its demonstrated effectiveness and efficiency in textual mathematical reasoning~\citep{deepseekmath,deepseek-r1}. Furthermore, we optimize the GRPO objective function by introducing dynamic hyper-parameter adjustment, which leads to improved model inference performance.

Mathematically, let $q$ denote a query and $\{o_i\}_{i=1}^G$ represent a set of $G$ completions generated by the old policy model $\pi_{old}$. Each completion $o_i$ is assigned a reward $R_i$ based on the task type of $q$, and the corresponding advantage estimate is formulated as:
\begin{equation}
    \hat{A}_{i} = \frac{R_i - \text{mean}\left( \{ R_i \}_{i=1}^G\right)}{\text{std}\left(\{ R_i \}_{i=1}^G\right)}
\end{equation}

The reward $R_i$ for each completion $o_i$ consists of two components: the accuracy reward $R_{i}^{\text{acc}}$, which measures how correct the generated answer is, and the format reward $R_{i}^{\text{fmt}}$, which evaluates whether the response adheres to the required output structure. The \textbf{accuracy reward} $R_{i}^{\text{acc}}$ is computed in a rule-based manner, with the specific calculation method determined by the task type (general or spatial). A detailed description of the computation can be found in Section~\ref{sec:task-specific-accuracy-reward-formulation}. The \textbf{format reward} $R_{i}^{\text{fmt}}$, on the other hand, checks whether the response includes valid thinking and answer blocks, i.e. \texttt{<think></think>} and \texttt{<answer></answer>}. It is defined as:

\begin{equation}
    R^{\text{fmt}}_i = \mathbb{I}(\text{Valid thinking and answer blocks in }o_i)
\end{equation}
where the indicator function $\mathbb{I}$ returns 1 if the formatting is valid and 0 otherwise. The final reward can be computed as:
\begin{equation}
    R_i = R_{i}^{\text{acc}} + R_{i}^{\text{fmt}}.
\end{equation}

Our variant of GRPO maximizes the following objective function:
\begin{equation}
\label{eq:grpo}
\begin{split}
    &\mathcal{J}_{GRPO} = \mathbb{E}_{q, \{o_i\} \sim \pi_{\theta_{old}}} \\
    &\left[\frac{\textcolor{blue}{\hat{\alpha}_{\{o_i\}}}}{G} \sum_{i=1}^{G} \frac{1}{\vert o_i \vert} \sum_{t=1}^{\vert o_i \vert} \min \left( r_{i, t}\left(\theta \right) {\hat{A}_{i, t}}, \text{clip}\left(r_{i, t}\left(\theta \right), 1-\epsilon, 1+\epsilon\right) {\hat{A}_{i, t}} \right) - \textcolor{blue}{\hat{\beta}} \mathbb{D}_{KL}\left(\pi_{\theta} \Vert \pi_{\text{ref}}\right) \right]
\end{split}
\end{equation}
where $r_{i, t}\left(\theta \right)$ represents the importance sampling term and is computed as:
\begin{equation}
    r_{i, t}\left(\theta \right) = \frac{\pi_{\theta}\left(o_{i, t} | q, o_{i,<t}\right)}{\pi_{\theta_{old}}\left(o_{i, t} | q, o_{i,<t}\right)}
\end{equation}

The \textcolor{blue}{highlighted terms} in Eq.~\ref{eq:grpo} represent our key modification—the introduction of dynamic hyper-parameter adjustment—distinguishing our method from the vanilla GRPO. These extensions are further elaborated in Section~\ref{sec:step-wise-dynamic-optim}.

\subsubsection{Curriculum Sampling}
\label{sec:curriculum-sampling}
Motivated by the progressive nature of human learning, we design our training framework to emulate the way humans acquire knowledge—starting with simple concepts and gradually advancing to more complex and challenging tasks. This staged learning process mirrors a curriculum-like structure, where each level builds upon the previous one to foster deeper understanding and reasoning.

Several recent studies~\citep{Kimi-k1.5,kimi-vl,Observe-R1} have highlighted the effectiveness of curriculum-based sampling strategies in improving model generalization and learning efficiency. Following this principle, we \textit{offline} pre-score and filter the training data based on task difficulty prior to the RLVR stage. The selected samples are then organized in increasing order of difficulty, enabling a step-wise, difficulty-ascending training paradigm that supports stable and effective learning.

\subsubsection{Step-wise Dynamic Optimization}
\label{sec:step-wise-dynamic-optim}

Our joint training framework is designed to overcome two primary obstacles to efficient learning. First, we address the pronounced data heterogeneity among tasks with differing modalities and complexities. To harmonize the training process, we employ a balanced data sampling mechanism that prevents any single task from dominating the learning process. Second, we recognize that a one-size-fits-all training approach is suboptimal for handling data of varying difficulty. To address this, we introduce dynamic hyper-parameter adjustment to focus the model's effort on more instructive samples, thereby optimizing the learning process.

\textbf{Data Sampling Balance.} Following~\citep{guo2025m2}, we adopt two principles for data sampling during both the cold-start and RLVR stages: (1) each training batch contains data from only one specific task, and (2) training steps are uniformly distributed across all tasks throughout the training pipeline. These strategies ensure balanced computational resource allocation while preserving task-specific learning characteristics.

\textbf{Dynamic Hyper-parameter Adjustment.} During the RLVR stage, we introduce several modifications to the standard GRPO algorithm to improve performance, including dynamic advantage weighting and a cosine-annealed KL penalty, as described in Section~\ref{sec:two-stage-training-strategy}.

Specifically, for advantage weighting, we adopt the insight from \citep{Observe-R1} that data of moderate difficulty, which corresponding to an \textit{online} accuracy reward of approximately 0.5, contains more informative signals. Such samples are therefore assigned higher weights. This leads to the following weight calculation:
\begin{equation}
    \hat{\alpha}_{{\{o_i\}}_{i=1}^G} = \sigma \text{mean}\left(\{R^{\text{acc}}_{i}\}_{i=1}^G\right)\left(1 - \text{mean}\left(\{R^{\text{acc}}_{i}\}_{i=1}^G\right)\right)
\end{equation}
where the scaling factor $\sigma$ is set to 7.2. It is crucial to note that our dynamic weighting and curriculum sampling are two orthogonal components: curriculum sampling provides a static, pre-determined data ordering based on intrinsic difficulty, while dynamic weighting modulates sample importance in real-time, adapting to the model's evolving capabilities.

Second, we employ a cosine annealing strategy for the KL penalty coefficient~\citep{Visionary-R1}. Combined with our curriculum sampling approach, this helps stabilize training on easier data while encouraging longer and more meaningful outputs for more challenging examples. The updated coefficient $\hat{\beta}$ is computed as:
\begin{equation}
    \hat{\beta} = \frac{\beta}{2}\left(1+\cos{\left(\pi\frac{T_{\text{cur}}}{T_{\text{max}}}\right)}\right)
\end{equation}
where $T_{\text{cur}}$ and $T_{\text{max}}$ denote the current and maximum training steps, respectively.


\subsection{Task-specific Accuracy Reward Formulation}
\label{sec:task-specific-accuracy-reward-formulation}
In this section, we provide a comprehensive explanation of the accuracy reward computation methods across various reasoning tasks, detailing the evaluation criteria and implementation strategies for each task type.

\textbf{General Reasoning.}
The general reasoning data comprises two question types: multiple-choice questions and fill-in-the-blank questions. To effectively evaluate model outputs across these diverse formats, we design a rule-based reward mechanism that accounts for four distinct response types — exact option letter matching, numeric digit matching, extract string matching, and formal mathematical expression verification. Toward this end, we integrate the Math-Verify\footnote{\url{https://github.com/huggingface/Math-Verify}} package and customized string matching algorithm with case and punctuation insensitivity to rigorously assess the accuracy of generated responses.

Accordingly, the accuracy reward for the general reasoning task is defined as:
\begin{equation}
    R^{\text{gen}}_i = \mathbb{I}\left(o_i, \text{ground truth}\right)
\end{equation}
where $\mathbb{I}$ denotes an indicator function that returns 1 if $o_i$ matches the ground truth, and 0 otherwise.

\textbf{Spatial Reasoning.} In spatial reasoning tasks, certain fill-in-the-blank questions demand specific numerical answers—such as size of objects and distances between objects—where an exact match may not always be appropriate. Current MLLMs often struggle to percept absolute spatial metrics, frequently producing rough and inaccurate estimates, particularly in the early stages of training. When binary rewards are assigned solely based on exact matches, the model faces significant challenges in generating responses that qualify as positive samples. This limitation hampers gradient accumulation impeding the learning process. Consequently, a more nuanced reward function is essential to better assess model outputs and facilitate meaningful learning, as it can provide a smooth reward even when the model's initial predictions deviate significantly from the actual values, consistently encouraging it to optimize in the correct direction.

\begin{wrapfigure}{r}{0.5\textwidth}
  \centering
  \includegraphics[width=1.0\linewidth]{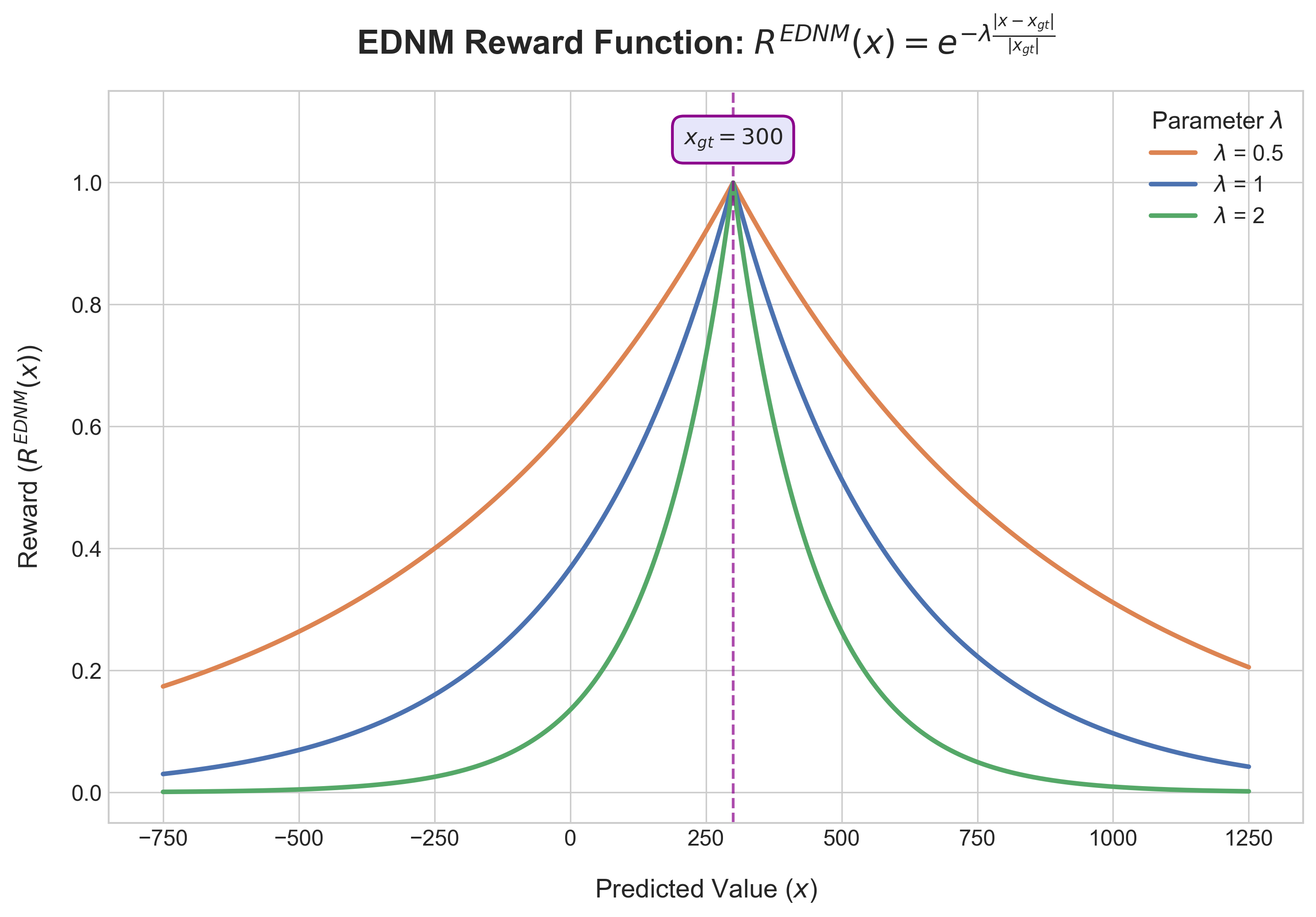}
  \caption{Visualization of the EDNM reward function for different values of the hyperparameter $\lambda$.}
\label{fig:ednm_distribution}
\end{wrapfigure}
We designed a reward calculation mechanism, which we term Exponential Decay Numeric Matching (EDNM), based on smooth continuous absolute values. The core of this mechanism is a continuous reward function that penalizes predictions based on their normalized relative error. The reward $R^{\text{EDNM}}(x)$ is calculated as follows:
\begin{equation}
    R^{\text{EDNM}}(x) = \gamma \cdot \exp\left(-\lambda \cdot \frac{|x - x_{\text{gt}}|}{|x_{\text{gt}}| + \epsilon}\right)
\end{equation}
where $\gamma$ and $\lambda$ are tunable hyper-parameters. Specifically, $\gamma$ serves as a scaling factor defining the maximum reward, while $\lambda$ controls the decay rate, determining the function's sensitivity to prediction errors. Figure~\ref{fig:ednm_distribution} illustrates this behavior, showing the reward for various predictions under different $\lambda$ settings, given a ground truth of $x_{\text{gt}} = 300$. For our experiments, we set the hyper-parameters to $\gamma = 1$ and $\lambda = 2$.

\section{Experiments}\label{sec:experiments}
We conduct a comprehensive evaluation of our models across two key domains: general and spatial reasoning. Our evaluation utilizes a diverse set of public benchmarks, grouped by the primary capability they measure:

\begin{enumerate}[label=$\blacklozenge$, wide=0pt, leftmargin=*]
    \item \textbf{General Reasoning (Mathematical \& Logical, Section~\ref{sec:general_reasoning_evaluation}):} To evaluate this capability, we employ six benchmarks: MathVista~\citep{mathvista}, MathVision~\citep{mathvision}, MathVerse~\citep{mathverse}, DynaMath~\citep{dynamath}, WeMath~\citep{wemath}, and LogicVista~\citep{logicvista}.
    \item \textbf{Spatial Reasoning (Section~\ref{sec:spatial_reasoning_evaluation}):} We assess this skill using 2 benchmarks: CV-Bench~\citep{cv-bench} and VSI-Bench~\citep{vsi-bench}.
\end{enumerate}

\subsection{General Reasoning Evaluation}
\label{sec:general_reasoning_evaluation}

As presented in Table~\ref{tab:math_benchmark}, we evaluate our \method-7B against other leading base-scale MLLMs on general multimodal reasoning benchmarks. Our model achieves a new state-of-the-art average score of 45.0, surpassing both top general models like InternVL3-8B (41.4) and reasoning models such as WeThink-VL-7B (44.3). Notably, \method-7B obtains the best results on MathVista (75.0) and DynaMath (26.8), and secures competitive second-place rankings on MathVision (31.5) and LogicVista (50.0). This performance, representing a significant \textbf{+9.5} point improvement over our base model, underscores the effectiveness of our approach in enhancing multimodal reasoning capabilities.

\begin{table}[t]
  \centering
    \caption{
        \textbf{Performance of MLLMs on general multimodal reasoning benchmarks.} 
        Best and second-best results of base-scale MLLMs are in \textbf{bold} and \underline{underlined}, respectively. 
        $\Delta$ denotes the performance improvement over the corresponding base model. 
        All scores are from the OpenCompass leaderboard~\citep{vlmevalkit}.
    }
  \label{tab:math_benchmark}
  \resizebox{0.9\textwidth}{!}{
  \begin{tabular}{l|cccccc|l}
    \toprule
    & Math- & Math- & Math- & Dyna- & WeMath & Logic- &  \\
    Models & Vista & Vision & Verse & Math & & Vista & Avg. ($\Delta$) \\
    \midrule
    \multicolumn{8}{l}{\textbf{\textit{Base-Scale General Models}}} \\
    InternVL3-8B~\citep{zhu2025internvl3}         & 70.5 & 30.0 & 38.5 & \underline{25.7} & 39.5 & 44.5 & 41.4 \\
    InternVL3-9B~\citep{zhu2025internvl3}         & 69.0 & 29.3 & 37.9 & 25.1 & 34.8 & 49.0 & 40.8 \\
    Qwen2.5-VL-7B~\citep{qwen2.5-vl}        & 68.1 & 25.4 & 41.1 & 21.8 & 36.2 & 47.9 & 40.1 \\
    MUG-U-7B~\citep{mug-u}             & \underline{74.8} & 26.1 & 35.4 & 17.2 & 26.5 & 39.8 & 36.6 \\
    SAIL-VL-1.6-8B~\citep{sail-vl-1.6}       & 74.2 & 23.2 & 33.4 & 14.0 & 29.6 & 41.4 & 36.0 \\
    \midrule
    \multicolumn{8}{l}{\textbf{\textit{Base-Scale Reasoning Models}}} \\
    WeThink-VL-7B~\citep{wethink}        & 71.6 & 26.0 & 44.2 & 24.8 & \textbf{48.0} & \textbf{51.2} & \underline{44.3} (+4.2) \\
    Taichu-VLR-7B~\citep{taichu-vlr}        & 72.3 & 27.1 & \underline{46.7} & 23.0 & \underline{44.0} & 48.3 & 43.6 \\
    VLAA-Thinker-7B~\citep{vlaa-thinker}      & 68.0 & 26.4 & \textbf{48.2} & 22.4 & 41.5 & 48.5 & 42.5 (+2.4) \\
    URSA-8B-PS-GRPO~\citep{luo2025ursa}      & 67.8 & \textbf{31.8} & 41.5 & 22.4 & 38.3 & 44.7 & 41.1 (+8.2) \\
    Ovis2-8B~\citep{lu2024ovis}             & 71.8 & 25.9 & 42.3 & 20.4 & 27.2 & 39.4 & 37.8 \\
    \midrule
    \multicolumn{8}{l}{\textbf{\textit{Our Models}}} \\
    Base Model           & 70.2 & 25.9 & 30.5 & 20.2 & 27.2 & 37.8 & 35.5 \\
    \method-CI-7B           & 71.7 & 29.2 & 42.1 & 25.0 & 42.8 & 46.8 & 42.9 (+7.4) \\
    \cellcolor{lightblue}\method-7B & \cellcolor{lightblue}\textbf{75.0}& \cellcolor{lightblue}\underline{31.5}& \cellcolor{lightblue}44.7& \cellcolor{lightblue}\textbf{26.8}& \cellcolor{lightblue}41.8& \cellcolor{lightblue}\underline{50.0}& \cellcolor{lightblue}\textbf{45.0} (\textcolor{red}{+9.5}) \\
    \bottomrule
  \end{tabular}
}
\end{table}

\begin{table}[h]
  \centering
    \caption{
        \textbf{Performance of MLLMs on CV-Bench.} 
        Best and second-best scores of base-scale MLLMs for each metric are shown in \textbf{bold} and \underline{underlined}, respectively. 
        \textsuperscript{\dag}Scores adapted from~\citep{ray2025satdynamicspatialaptitude}; 
        \textsuperscript{*}scores reproduced by us.
    }
  \label{tab:cv-bench-performance}
  \resizebox{0.75\textwidth}{!}{%
  \begin{tabular}{l|cccc|c}
    \toprule
    Models & Count & Relation & Depth & Distance & Avg. \\
    \midrule
    \multicolumn{6}{l}{\textbf{\textit{Large-Scale Models}}} \\
    GPT-4O\textsuperscript{\dag}~\citep{achiam2023gpt}               & 65.9  & 85.7  & 87.8  & 78.2  & 78.9 \\
    Gemini-1.5-pro\textsuperscript{\dag}~\citep{team2023gemini}        & 70.4  & 85.2  & 82.4  & 72.8  & 77.4 \\
    
    \midrule
    \multicolumn{6}{l}{\textbf{\textit{Base-Scale Models}}} \\
    InternVL3-8B\textsuperscript{*}~\citep{zhu2025internvl3}         & \textbf{74.0}  & \underline{90.6}  & \underline{84.3}  & \underline{81.0}  & \underline{82.0} \\
    Qwen2.5-VL-7B-Instruct~\citep{qwen2.5-vl} & 65.2  & 86.6  & 70.6  & 79.8  & 75.0 \\
    LLava-NEXT-Video-7B\textsuperscript{\dag}~\citep{zhang2024llavanextvideo}         & 59.3  & 77.0  & 71.3  & 54.7  & 65.2 \\
    \midrule
    \multicolumn{6}{l}{\textbf{\textit{Our Models}}} \\
    \cellcolor{lightblue}\method-7B & \cellcolor{lightblue}\underline{66.6}& \cellcolor{lightblue}\textbf{92.8}& \cellcolor{lightblue}\textbf{89.3}& \cellcolor{lightblue}\textbf{84.3}& \cellcolor{lightblue}\textbf{82.3} \\
    \bottomrule
  \end{tabular}
  } 
\end{table}

\begin{table}[h]
  \centering
    \caption{
        \textbf{Performance of MLLMs on VSI-Bench.} 
        Best and second-best scores of base-scale MLLMs are shown in \textbf{bold} and \underline{underlined}.
        Abbreviations stand for: \texttt{OC} (Object Count), \texttt{AD} (Absolute Distance), \texttt{OS} (Object Size), \texttt{RS} (Room Size), \texttt{RDs} (Relative Distance), \texttt{RDr} (Relative Direction), \texttt{RP} (Route Plan), and \texttt{AO} (Appearance Order).
        \textsuperscript{\dag}Scores adapted from the InternVL3 technical report~\citep{zhu2025internvl3}.
    }
  \label{tab:vsi-bench-performance}
  \resizebox{0.9\textwidth}{!}{%
  \begin{tabular}{l|cccccccc|c}
    \toprule
    Models & \texttt{OC} & \texttt{AD} & \texttt{OS} & \texttt{RS} & \texttt{RDs} & \texttt{RDr} & \texttt{RP} & \texttt{AO} & Avg. \\
    \midrule
    \multicolumn{6}{l}{\textbf{\textit{Large-Scale Models}}} \\
    Gemini-1.5-pro\textsuperscript{\dag}~\citep{team2023gemini} & 56.2 & 30.9 & 64.1 & 43.6 & 51.3 & 46.3 & 36.0 & 34.6 & 45.4 \\
    GPT-4O\textsuperscript{\dag}~\citep{achiam2023gpt} & 46.2 & 5.3 & 43.8 & 38.2 & 37.0 & 41.3 & 31.5 & 28.5 & 34.0 \\
    \midrule
    \multicolumn{6}{l}{\textbf{\textit{Base-Scale Models}}} \\
    InternVL3-8B\textsuperscript{\dag}~\citep{zhu2025internvl3} & \textbf{68.1} & \textbf{39.0} & 48.4 & 33.6 & \textbf{48.3} & 36.4 & 27.3 & \textbf{35.4} & \underline{42.1} \\
    Video-R1-7B~\citep{video-r1} & - & - & - & - & - & - & - & - & 37.1 \\
    Qwen2.5-VL-7B-Instruct~\citep{qwen2.5-vl} & 37.7 & 20.1 & \underline{49.7} & \underline{37.4} & 38.5 & 40.4 & \underline{31.4} & \underline{32.0} & 35.9 \\
    LLava-NeXT-Video-7B\textsuperscript{\dag}~\citep{zhang2024llavanextvideo} & \underline{48.5} & 14.0 & 47.8 & 24.2 & \underline{43.5} & \underline{42.4} & \textbf{34.0} & 30.6 & 35.6 \\
    \midrule
    \multicolumn{10}{l}{\textbf{\textit{Our Models}}} \\
    \cellcolor{lightblue}\method-7B & \cellcolor{lightblue}41.0& \cellcolor{lightblue}\underline{34.0}& \cellcolor{lightblue}\textbf{60.9}& \cellcolor{lightblue}\textbf{55.4}& \cellcolor{lightblue}40.7& \cellcolor{lightblue}\textbf{47.3}& \cellcolor{lightblue}29.9& \cellcolor{lightblue}28.8& \cellcolor{lightblue}\textbf{42.3} \\
    \bottomrule
  \end{tabular}
  } 
\end{table}

\subsection{Spatial Reasoning Evaluation}
\label{sec:spatial_reasoning_evaluation}

As shown in Table~\ref{tab:cv-bench-performance}, our model, \method-7B, achieves state-of-the-art performance on CV-Bench with an average score of 82.3, slightly surpassing the previous leading model, InternVL3-8B (82.0). Our model demonstrates exceptional strength in understanding complex spatial configurations, securing the top scores in Relation (92.8), Depth (89.3), and Distance (84.3). While InternVL3-8B maintains an edge in object counting, \method-7B's superior ability to process and reason about relational and depth-related cues highlights its advanced spatial understanding capabilities.

On the more challenging VSI-Bench for nuanced video spatial imagination, \method-7B demonstrates highly competitive performance (Table~\ref{tab:vsi-bench-performance}). Its average score of 42.3 surpasses the strong InternVL3-8B baseline (42.1) and is second only to Gemini-1.5-pro (45.4). Notably, our model's key advantages lie in fine-grained analysis, where it establishes new state-of-the-art records for inferring Room Size (RS) at 55.4 and determining Relative Direction (RDr) at 47.3. Furthermore, it secures the second-best performance in estimating Object Size (OS) and Absolute Distance (AD). This highlights a sophisticated capability to infer complex spatial attributes like scale and orientation from dynamic scenes, showcasing its particular strengths in these nuanced reasoning tasks.

\label{sec:gui_reasoning_evaluation}
\section{Conclusion}\label{sec:conclusion}

In this paper, we introduce \method-7B, a model engineered to master both general and spatial tasks to address the critical deficiency of MLLMs in dynamic spatial reasoning. Our success hinges on the synergy between two core innovations: a sophisticated data pipeline that generates and curates high-quality data with verifiable reasoning trajectories, and a dynamic training strategy that resolves learning conflicts through step-wise optimization and task-specific rewards. This powerful integration enables \method-7B to establish a new state-of-the-art across eight diverse benchmarks. However, we acknowledge several limitations that warrant further investigation:
\begin{enumerate}[label=$\blacklozenge$, wide=0pt, leftmargin=*]
    \item \textbf{Constrained Reasoning Depth:} Compared to specialized language-only reasoning models like DeepSeek-R1, \method-7B tends to produce shorter reasoning chains. This limits its ability to deconstruct and solve highly complex, multi-step problems that require deeper logical deliberation.
    \item \textbf{Pathological Repetition:} In some instances, the model exhibits pathological repetition, where it gets trapped in a loop, redundantly generating the same phrases or reasoning steps. This behavior indicates a potential instability in the generation process that can derail coherent thought progression.
    \item \textbf{Suboptimal Visual Perception:} Despite its strong reasoning capabilities, the model's visual perception is not infallible. We observed occasional weaknesses in perceiving fine-grained details, leading to imprecise descriptions or, in some cases, visual hallucinations where the model fabricates non-existent objects or attributes.
\end{enumerate}

We will continue to explore and address these challenges in our future work to further enhance the model's robustness and reasoning capabilities.

\section{Contributors}
\label{sec:contributors}
Authors are listed \textbf{alphabetically by the first name}.

\begin{multicols}{3}
Fudong Wang \\
Jiajia Liu \\ 
Jingdong Chen \\
Jun Zhou \\
Kaixiang Ji \\
Lixiang Ru \\
Qingpei Guo \\
Ruobing Zheng \\
Tianqi Li \\
Yi Yuan \\
Yifan Mao \\
Yuting Xiao \\
Ziping Ma \\
\end{multicols}

\bigskip

\bibliographystyle{assets/plainnat}
\bibliography{paper}

\newpage
\appendix

\DefineVerbatimEnvironment%
  {NormalFontVerbatim}{Verbatim}{fontfamily=\rmdefault, formatcom=\color{black}}

\newtcolorbox{examplebox}[1]{
    enhanced,
    breakable,
    skin=bicolor,              
    arc=3mm, 
    colback=lightblue!25!white,
    colframe=lightblue,
    boxrule=1.5pt,
    title=#1,
    fonttitle=\bfseries\color{black},
    colbacktitle=lightblue,
    coltitle=white,
}

\section{Prompts Used for High-Quality Data Construction Pipeline}
\label{app:prompts}

\begin{tcolorbox}[
    enhanced,
    breakable,
    skin=bicolor,              
    arc=3mm,                   
    boxrule=1.5pt,             
    colframe=lightblue,
    colback=white,             
    colbacktitle=lightblue,
    title=Prompt for scoring synthesized general CoT data,       
    fonttitle=\bfseries\color{black}, 
]
You are given a written mathematical solution. Please rate the quality of its reasoning process on a scale from 1 (lowest) to 5 (highest) based on the following three criteria:

\begin{enumerate}[leftmargin=*, label=\arabic*.]
    \item \textbf{Optimal Structural Organization:} Is the reasoning process clearly and logically structured? Does it break down complex steps with sufficient detail, while keeping straightforward steps concise? High-quality solutions adapt the level of detail (granularity) to the complexity of each step, ensuring that critical transitions receive appropriate elaboration.

    \item \textbf{Effective Cognitive Scaffolding:} Does the solution guide the reader’s understanding by gradually building up concepts and insights? Are key ideas introduced at the right time, and does the explanation bridge conceptual gaps thoughtfully, making the solution easier to follow and learn from?

    \item \textbf{Rigorous Verification:} Does the reasoning include frequent verification of intermediate results? Are assumptions explicitly checked, and is logical consistency maintained throughout the solution? A high-quality answer thoroughly validates its correctness at each stage.
\end{enumerate}

\medskip

\textbf{Scoring Rubric:}
\begin{itemize}[leftmargin=*]
    \item \textbf{5 — Excellent:} Exemplary reasoning with adaptive step granularity, strong pedagogical clarity, and thorough verification.
    \item \textbf{4 — Good:} Clear and mostly well-paced explanation with minor issues in elaboration or verification.
    \item \textbf{3 — Fair:} Reasonable logic but with noticeable issues in structure, clarity, or verification.
    \item \textbf{2 — Poor:} Disorganized or incomplete reasoning with major gaps in explanation or checks.
    \item \textbf{1 — Very Poor:} Lacks coherent structure, explanation, and verification; difficult to understand or trust.
\end{itemize}

\medskip

The given reasoning process is: \texttt{REASONING}

\medskip

You MUST output your score with one single number.
\end{tcolorbox}

\begin{tcolorbox}[
    enhanced,         
    skin=bicolor,              
    arc=3mm,                   
    boxrule=1.5pt,             
    colframe=lightblue,
    colback=white,             
    colbacktitle=lightblue,
    title=Prompt for scoring the difficulty of RLVR data,       
    fonttitle=\bfseries\color{black}, 
]

\#\#\# For multiple-choice questions:\\
According to the following question, please first conduct step by step reasoning, then answer the question and put the correct option letter, e.g., A, B, C, D, within \textbackslash boxed\{\}.\\
\medskip\\
\#\#\# For fill-in-the-blank questions:\\
According to the following question, please first conduct step by step reasoning, then answer the question and put the final answer within \textbackslash boxed\{\}.
\end{tcolorbox}

\begin{tcolorbox}[
    enhanced, 
    skin=bicolor,              
    arc=3mm,                   
    boxrule=1.5pt,             
    colframe=lightblue,
    colback=white,             
    colbacktitle=lightblue,
    title=Prompt for validating the spatial reasoning data,       
    fonttitle=\bfseries\color{black}, 
]
You are a logic consistency evaluator. Evaluate the Input by the Evaluation Rules and output the final response with Output Format.

\medskip

\#\#\# Input:
\begin{itemize}[leftmargin=*]
    \item Question: \texttt{QUESTION}
    \item Choice List: \texttt{ANSWER\_CHOICES}
    \item Answer: \texttt{CORRECT\_ANSWER}
\end{itemize}

\medskip

\#\#\# Evaluation Rules:
\begin{enumerate}[leftmargin=*]
    \item Thoroughly analyze the provided image, question, answer, and the choice list before scoring.
    \item Evaluate the data quality based on the following three dimensions, providing a score of 1 (Good) or 0 (Bad) for each.
    \begin{itemize}[leftmargin=*, label=\textendash]
        \item ObjectS (Object Score): Assess if the object(s) relevant to the question are clearly visible and identifiable in the image.
        \begin{itemize}[leftmargin=*, label=\textendash]
            \item Score 1 if the objects are clear and unambiguous.
            \item Score 0 if the objects are significantly occluded, blurry, overlapping, too small, cut off at the image edge, or otherwise difficult to discern in a way that impedes answering the question.
        \end{itemize}
        
        \item AnswerS (Answer Score): Verify if the provided answer is correct and unambiguous based on the image and the question.
        \begin{itemize}[leftmargin=*, label=\textendash]
            \item Score 1 if the answer is factually correct and logically sound.
            \item Score 0 if the answer is ambiguous, factually incorrect, inconsistent with the image, or cannot be confirmed from the provided information.
        \end{itemize}
        
        \item OptionS (Options Score): Examine the quality and validity of the Choice List.
        \begin{itemize}[leftmargin=*, label=\textendash]
            \item Score 1 if all choices are meaningful, distinct, and plausible distractors that don't confuse the question's intent.
            \item Score 0 if choices are redundant, highly similar, nonsensical, or clearly irrelevant to the question or image.
        \end{itemize}
    \end{itemize}
\end{enumerate}

\medskip

\#\#\# Output Format:
\begin{enumerate}[leftmargin=*]
    \item Output MUST be valid JSON format which can be loaded by json.loads() and contains the reasoning step.
    \item Example:
    \begin{NormalFontVerbatim}
{
    "Reason\_Step": <reasoning step by step>,
    "ObjectS": 1,
    "AnswerS": 1,
    "OptionS": 0
}
    \end{NormalFontVerbatim}
    \item Please reason step by step, and put your final JSON format output within \texttt{\textbackslash boxed\{\}}.
\end{enumerate}
\end{tcolorbox}

\section{Experimental Implementation Details}

\subsection{System Prompts}

To guide the reasoning process during training, we employ the following system prompt in both the cold-start and RLVR stages:

\begin{tcolorbox}[
    enhanced,                  
    skin=bicolor,              
    arc=3mm,                   
    boxrule=1.5pt,             
    colframe=lightblue,
    colback=white,             
    colbacktitle=lightblue,
    title=System Prompt,       
    fonttitle=\bfseries\color{black},
]
You are a helpful assistant. When the user asks a question, your response must include two parts: first, the reasoning process enclosed in <think>...</think> tags, then the final answer enclosed in <answer>...</answer> tags. The critical answer or key result should be placed within \textbackslash boxed\{\}.
\end{tcolorbox}

\subsection{Training Hyper-parameters for RLVR}

Our training framework is built upon VisualThinker-R1-Zero~\citep{visualthinker}, and all RLVR hyper-parameters are presented in Table~\ref{tab:all_hyper_parameters}.

\begin{table}[htbp]
\centering
\caption{Training Hyper-parameters for RLVR}
\label{tab:all_hyper_parameters}
\begin{tabular}{lccc}
\toprule
\textbf{} & \textbf{General} & \textbf{Spatial\textsubscript{image}} & \textbf{Spatial\textsubscript{video}} \\
\midrule
Learning Rate & \multicolumn{3}{c}{\texttt{1e-6}} \\
Initial KL Coefficient & \multicolumn{3}{c}{\texttt{0.01}} \\
Completion Max Length & \multicolumn{3}{c}{\texttt{1024}} \\
Temperature & \multicolumn{3}{c}{\texttt{1.0}} \\
Batch Size & \multicolumn{3}{c}{\texttt{64}} \\
Epochs & \texttt{1} & \texttt{2} & \texttt{3} \\
Number of Generations($G$) & \texttt{16} & \texttt{16} & \texttt{4} \\
\bottomrule
\end{tabular}
\end{table}

\subsection{Benchmarks}
As illustrated in Section~\ref{sec:general_reasoning_evaluation}, we conduct evaluation across six mathematical and logical reasoning benchmarks. Below are the details:

\begin{enumerate}[label=$\lozenge$, wide=0pt, leftmargin=*]
    \item \textbf{MathVista:} We use the Test Mini split of the MathVista dataset.
    \item \textbf{MathVision:} We use the full test set of the MathVision dataset.
    \item \textbf{MathVerse:} We use the Test Mini split of MathVerse in "Vision Only" mode.
    \item \textbf{DynaMath:} We use the full test set of DynaMath, and report the worst-case accuracy as the main metric.
    \item \textbf{WeMath:} We use the Test Mini split of WeMath, and report the "Score (Strict)" as the main metric.
    \item \textbf{LogicVista:} We use the full test set of LogicVista.
\end{enumerate}

\section{Visulization Results}

The following examples showcases visualization results of \method on general and spatial reasoning tasks. These examples demonstrate that \method-7B possesses strong logical and analytical capabilities, enabling it to arrive at the correct answers.

\begin{examplebox}{General Reasoning}
    \includegraphics[width=0.5\linewidth]{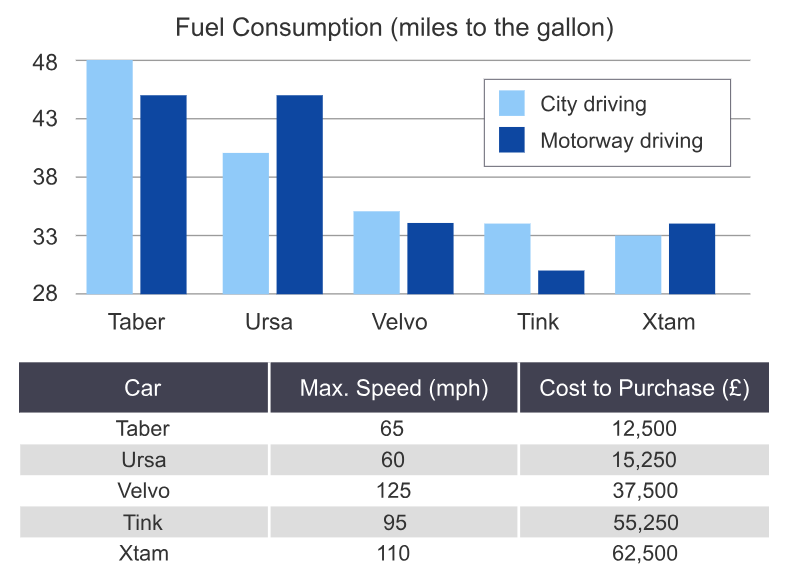} 

    \textbf{Question:} If a driver travels an average of 4,250 miles per month driving only along motorways in an Xtam car, what is the predicted annual consumption of fuel (in gallons)? Select from A, B, C, D and E.\\
    \textbf{Options:} \\
    (A) Cannot say \\
    (B) 1,500 \\
    (C) 125 \\
    (D) 150 \\
    (E) 1,250 \\
    \textbf{GT Answer:} B

    \bigskip
    \hrule
    \bigskip

    \textbf{\method-7B:} \\
    \smallskip
    
    <think>\\
    To determine the predicted annual consumption of fuel for an Xtam car driven only along motorways, we need to follow these steps:\\
    \\
    1. Identify the fuel consumption for the Xtam car on the motorway from the bar chart.\\
    2. Calculate the monthly fuel consumption.\\
    3. Multiply the monthly fuel consumption by 12 to get the annual consumption.\\
    \\
    From the bar chart:\\
    - The fuel consumption for the Xtam car on the motorway is 33.5 miles per gallon.\\
    \\
    Next, we need to calculate the monthly fuel consumption:\\
    - The driver travels 4,250 miles per month.\\
    - Fuel consumption = Total miles driven / Miles per gallon\\
    - Monthly fuel consumption = 4,250 miles / 33.5 miles per gallon $\approx$ 126.76 gallons\\
    \\
    Finally, we calculate the annual fuel consumption:\\
    - Annual fuel consumption = Monthly fuel consumption * 12\\
    - Annual fuel consumption $\approx$ 126.76 gallons/month * 12 months $\approx$ 1,521.12 gallons\\
    \\
    The closest option to 1,521.12 gallons is 1,500 gallons.\\
    \\
    Therefore, the predicted annual consumption of fuel is approximately 1,500 gallons.\\
    </think>\\
    <answer>\\
    \boxed{B}\\
    </answer>
\end{examplebox}

\begin{examplebox}{General Reasoning}
    \includegraphics[width=0.5\linewidth]{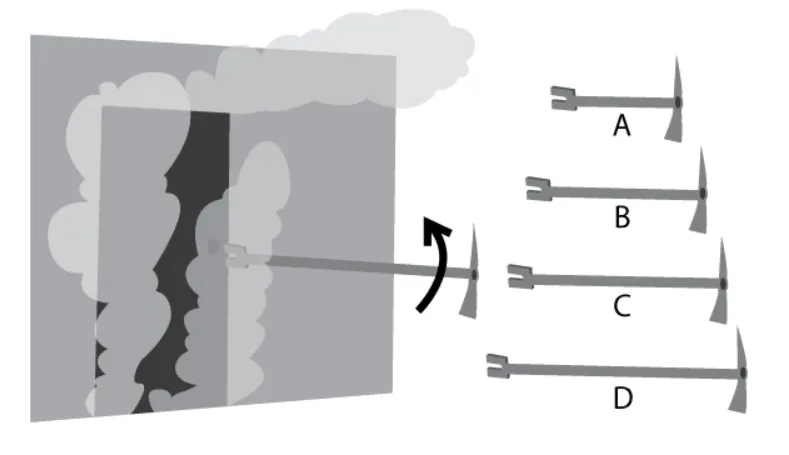} 

    \textbf{Question:} Firefighters use a halligan to forcibly enter locked doors during fire rescue operations. Which halligan would require the least effort to forcibly open the door? Select from A, B, C, D, and E.\\
    \textbf{Options:} \\
    (A) A \\
    (B) B \\
    (C) C \\
    (D) D \\
    (E) They would all require the same effort \\
    \textbf{GT Answer:} D

    \bigskip
    \hrule
    \bigskip

    \textbf{\method-7B:} \\
    \smallskip
    
    <think>\\
    To determine which Halligan tool would require the least effort to forcibly open the door, we need to consider the mechanical advantage provided by each tool. The Halligan tool is a lever, and the effort required to lift the tool is inversely proportional to the distance from the fulcrum to the point where the force is applied.\\
    \\
    - Tool A has a short handle and a long lever arm.\\
    - Tool B has a slightly longer handle and a shorter lever arm.\\
    - Tool C has a medium-length handle and a medium-length lever arm.\\
    - Tool D has the longest handle and the shortest lever arm.\\
    \\
    The longer the handle, the greater the mechanical advantage, meaning less effort is required to lift the tool. Therefore, the tool with the longest handle and the shortest lever arm would require the least effort.\\
    \\
    From the options given, Tool D has the longest handle and the shortest lever arm, providing the greatest mechanical advantage.\\
    \\
    Thus, the tool that would require the least effort to forcibly open the door is Tool D.\\
    </think>\\
    <answer>\\
    \boxed{D}\\
    </answer>
\end{examplebox}

\begin{examplebox}{General Reasoning}
    \includegraphics[width=0.5\linewidth]{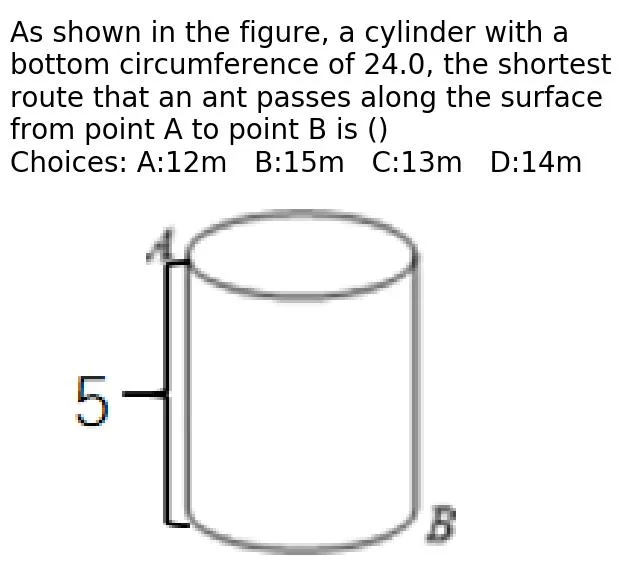} 

    \textbf{Question:} According to the question shown in the image, please directly answer the question and provide the correct option letter, e.g., A, B, C, D.\\
    \textbf{GT Answer:} C

    \bigskip
    \hrule
    \bigskip

    \textbf{\method-7B:} \\
    \smallskip
    
    <think>\\
    To solve this problem, we need to determine the shortest path an ant would take on the surface of the cylinder from point A to point B. The key is to \"unroll\" the cylinder into a flat surface, which will help us visualize the path more easily.\\
    \\
    1. \textbf{Circumference of the Cylinder}: The circumference is given as 24.0 meters. This means if we unroll the cylinder, the length of the bottom edge of the rectangle will be 24.0 meters.\\
    2. \textbf{Height of the Cylinder}: The height of the cylinder is given as 5 meters.\\
    3. \textbf{Unrolling the Cylinder}: When we unroll the cylinder, it forms a rectangle where one side is the height of the cylinder (5 meters) and the other side is the circumference of the base (24.0 meters).\\
    4. \textbf{Points A and B}: Point A is at the bottom left corner of the rectangle, and point B is at the top right corner of the rectangle.\\
    5. \textbf{Shortest Path}: The shortest path from A to B on the unrolled rectangle is the straight-line distance between these two points. This forms a right triangle where one leg is the height of the cylinder (5 meters) and the other leg is half the circumference (since the ant can go either directly around the cylinder or the other way, but the shortest path is the direct line across the rectangle).\\
    \\
    The distance can be calculated using the Pythagorean theorem:\\
    $ \text{Distance} = \sqrt{(24.0)^2 + (5)^2} = \sqrt{576 + 25} = \sqrt{601} \approx 24.5 \text{ meters} $ \\
    However, the options provided are much smaller, suggesting we might need to consider the direct path around the cylinder or a simpler calculation. Given the options, the correct path is likely the direct path across the rectangle, which is the hypotenuse of a right triangle with legs 5 meters and 12 meters (half the circumference).\\
    $ \text{Distance} = \sqrt{(12)^2 + (5)^2} = \sqrt{144 + 25} = \sqrt{169} = 13 \text{ meters} $ \\
    Thus, the shortest path is 13 meters.\\
    </think>\\
    <answer>\\
    \boxed{C}\\
    </answer>
\end{examplebox}

\begin{examplebox}{General Reasoning}
    \includegraphics[width=0.4\linewidth]{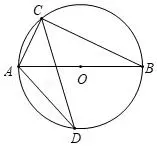} 

    \textbf{Question:} As shown in the figure, AB is the diameter of $\odot$O, CD is the chord of $\odot$O, $\angle$ADC = 26.0, then the degree of $\angle$ CAB is ()\\
    \textbf{Options:} \\
    (A) 26$^\circ$\\
    (B) 74$^\circ$\\
    (C) 64$^\circ$\\
    (D) 54$^\circ$\\
    \textbf{GT Answer:} C

    \bigskip
    \hrule
    \bigskip

    \textbf{\method-7B:} \\
    \smallskip
    
    <think>\\
    To solve the problem, we need to use some properties of circles and angles.\\
    \\
    1. Since AB is the diameter of the circle, the angle subtended by the diameter at any point on the circle is a right angle. This means $\angle$ACB = 90$^\circ$.\\
    2. We are given that $\angle$ADC = 26$^\circ$. Since $\angle$ADC and $\angle$ABC are both inscribed angles that subtend the same arc AC, they are equal. Therefore, $\angle$ABC = 26$^\circ$.\\
    3. In triangle ABC, we know that the sum of the angles in a triangle is 180$^\circ$. So, we can write:\\
    $\angle CAB + \angle ABC + \angle ACB = 180^\circ$\\
    Substituting the known values:\\
    $\angle CAB + 26^\circ + 90^\circ = 180^\circ$\\
    Simplifying this, we get:\\
    $\angle CAB + 116^\circ = 180^\circ$\\
    $\angle CAB = 180^\circ - 116^\circ = 64^\circ$\\
    \\
    Therefore, the measure of $\angle$CAB is 64$^\circ$.\\
    </think>\\
    <answer>\\
    $\boxed{C) 64^\circ}$\\
    </answer>
\end{examplebox}

\begin{examplebox}{General Reasoning}
    \includegraphics[width=0.4\linewidth]{images/general_reasoning_5.png} 

    \textbf{Question:} The diagrams below show two pure samples of gas in identical closed, rigid containers. Each colored ball represents one gas particle. Both samples have the same number of particles. Compare the average kinetic energies of the particles in each sample. Which sample has the higher temperature?\\
    \textbf{Options:} \\
    (A) neither; the samples have the same temperature\\
    (B) sample A\\
    (C) sample B\\
    \textbf{GT Answer:} B

    \bigskip
    \hrule
    \bigskip

    \textbf{\method-7B:} \\
    \smallskip
    
    <think>\\
    To determine which sample has the higher temperature, we need to compare the average kinetic energies of the particles in each sample. The average kinetic energy of a gas particle is given by the formula:\\
    $\text{Kinetic Energy} = \frac{1}{2}mv^2 $\\
    where $m$ is the mass of the particle and $v$ is its speed.\\
    \\
    The temperature of a gas is directly proportional to the average kinetic energy of its particles. Therefore, the sample with the higher average kinetic energy will have the higher temperature.\\
    \\
    Let's calculate the average kinetic energy for each sample:\\
    \\
    For Sample A:\\
    - Mass of each particle, $ m_A = 40 \text{u} $\\
    - Average particle speed, $ v_A = 950 \text{m/s} $\\
    \\
    Kinetic Energy for Sample A:
    $ KE_A = \frac{1}{2} \times 40 \times (950)^2 $\\
    \\
    For Sample B:\\
    - Mass of each particle, $ m_B = 32 \text{u} $\\
    - Average particle speed, $ v_B = 750 \text{m/s} $\\
    \\
    Kinetic Energy for Sample B:\\
    $ KE_B = \frac{1}{2} \times 32 \times (750)^2 $\\
    \\
    Now, let's compare the two kinetic energies:\\
    $ KE_A = \frac{1}{2} \times 40 \times (950)^2 $\\
    $ KE_B = \frac{1}{2} \times 32 \times (750)^2 $\\
    \\
    Since the mass and speed of the particles in Sample A are both higher than those in Sample B, the kinetic energy of the particles in Sample A will be higher.\\
    \\
    Therefore, Sample A has a higher average kinetic energy, and thus a higher temperature.\\
    </think>\\
    <answer>\\
    $\boxed{B) sample A}$\\
    </answer>
\end{examplebox}

\begin{examplebox}{Spatial Reasoning}
    \includegraphics[width=0.5\linewidth]{images/spatial_image_depth.png} 

    \textbf{Question:} Which object is closer to the camera taking this photo, the chair (highlighted by a red box) or the bin (highlighted by a blue box)? \\
    \textbf{Options:} \\
    (A) chair \\
    (B) bin \\
    \textbf{GT Answer:} B

    \bigskip
    \hrule
    \bigskip

    \textbf{\method-7B:} \\
    \smallskip
    
    <think>\\
    To determine which object is closer to the camera, we need to analyze the relative positions and sizes of the objects in the image. The chair highlighted by the red box is positioned in the background, far from the camera, as it is located at the far end of the room. The bin highlighted by the blue box is closer to the foreground, as it is situated near the center of the image and appears larger in the frame, indicating it is nearer to the camera.\\
    \\
    Given the spatial relationships and the perspective, the bin is closer to the camera than the chair.\\
    </think>\\
    <answer>\\
    \boxed{(B) bin}\\
    </answer>

\end{examplebox}

\begin{examplebox}{Spatial Reasoning}
    \includegraphics[width=0.5\linewidth]{images/spatial_image_distance.png} 

    \textbf{Question:} Estimate the real-world distances between objects in this image. Which object is closer to the chair (highlighted by a red box), the bookcase (highlighted by a blue box) or the table (highlighted by a green box)? \\
    \textbf{Options:} \\
    (A) bookcase \\
    (B) table \\
    \textbf{GT Answer:} B

    \bigskip
    \hrule
    \bigskip

    \textbf{\method-7B:} \\
    \smallskip
    
    <think>\\
    To determine which object is closer to the chair highlighted by the red box, we need to consider the spatial relationships between the objects in the image.\\
    \\
    1. The chair is located on the right side of the image.\\
    2. The bookcase is on the left side of the image, behind the couch and the desk.\\
    3. The table is in the center of the image, closer to the chair than the bookcase.\\
    \\
    Given the positions:\\
    - The table is directly in front of the chair and appears to be closer to it.\\
    - The bookcase is further away, on the opposite side of the room.\\
    \\
    Therefore, the table is closer to the chair than the bookcase.\\
    </think>\\
    <answer>\\
    \boxed{(B) table}\\
    </answer>
\end{examplebox}

\begin{examplebox}{Spatial Reasoning}
    \includegraphics[width=0.35\linewidth]{images/spatial_image_relation.png} 

    \textbf{Question:} Considering the relative positions of the person and the dog in the image provided, where is the person located with respect to the dog? Select from the following choices. \\
    \textbf{Options:} \\
    (A) left\\
    (B) right\\
    \textbf{GT Answer:} A

    \bigskip
    \hrule
    \bigskip

    \textbf{\method-7B:} \\
    \smallskip
    
    <think>\\
    The image shows a dog wearing a cowboy hat sitting in the back seat of a car. To the left of the dog, there is a person wearing a plaid shirt. The dog is positioned more towards the right side of the image, and the person is to the left of the dog. Therefore, the person is located to the left of the dog.\\
    </think>\\
    <answer>\\
    \boxed{(A) left}\\
    </answer>
\end{examplebox}

\begin{examplebox}{Spatial Reasoning}
    \includegraphics[width=0.7\linewidth]{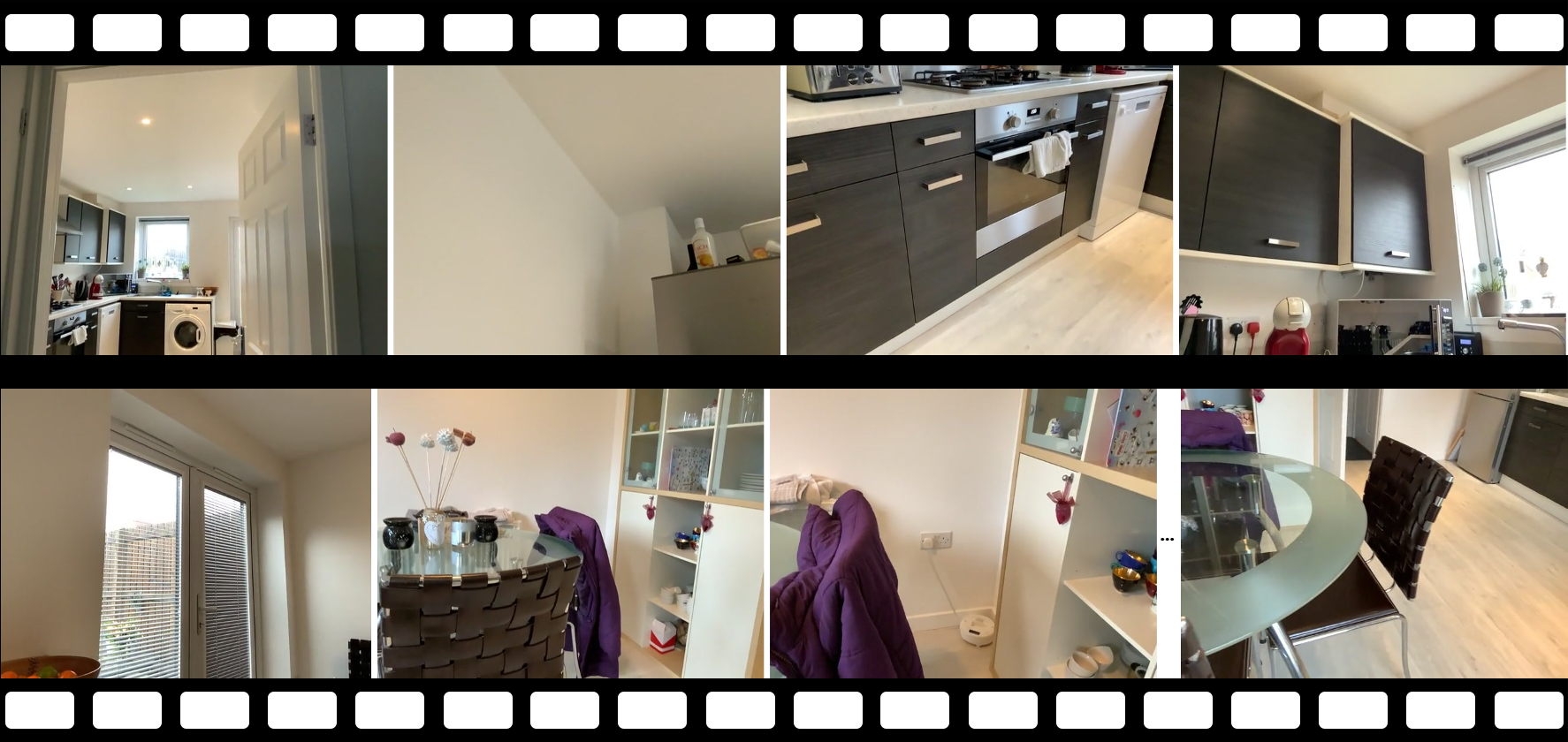} 

    \textbf{Question:} These are frames of a video. How many chair(s) are in this room? Please answer the question using a single word or phrase.\\
    \textbf{GT Answer:} 3

    \bigskip
    \hrule
    \bigskip

    \textbf{\method-7B:} \\
    \smallskip
    
    <think>\\
    To determine the number of chairs in the room, I will carefully analyze the images provided, focusing on identifying distinct chair shapes and structures. I will note any visible chairs and ensure no duplicates are counted.\\
    The images show a modern kitchen and a dining area. In the kitchen, there is a visible chair near the dining table. Another chair is seen in the dining area, and a third chair is partially visible in the background. No other chairs are evident in the other sections of the room.\\
    Counting the chairs: 1 (near the dining table) + 1 (in the dining area) + 1 (partially visible) = 3 chairs.\\
    </think>\\
    <answer>\\
    \boxed{3}\\
    </answer>
\end{examplebox}

\begin{examplebox}{Spatial Reasoning}
    \includegraphics[width=0.7\linewidth]{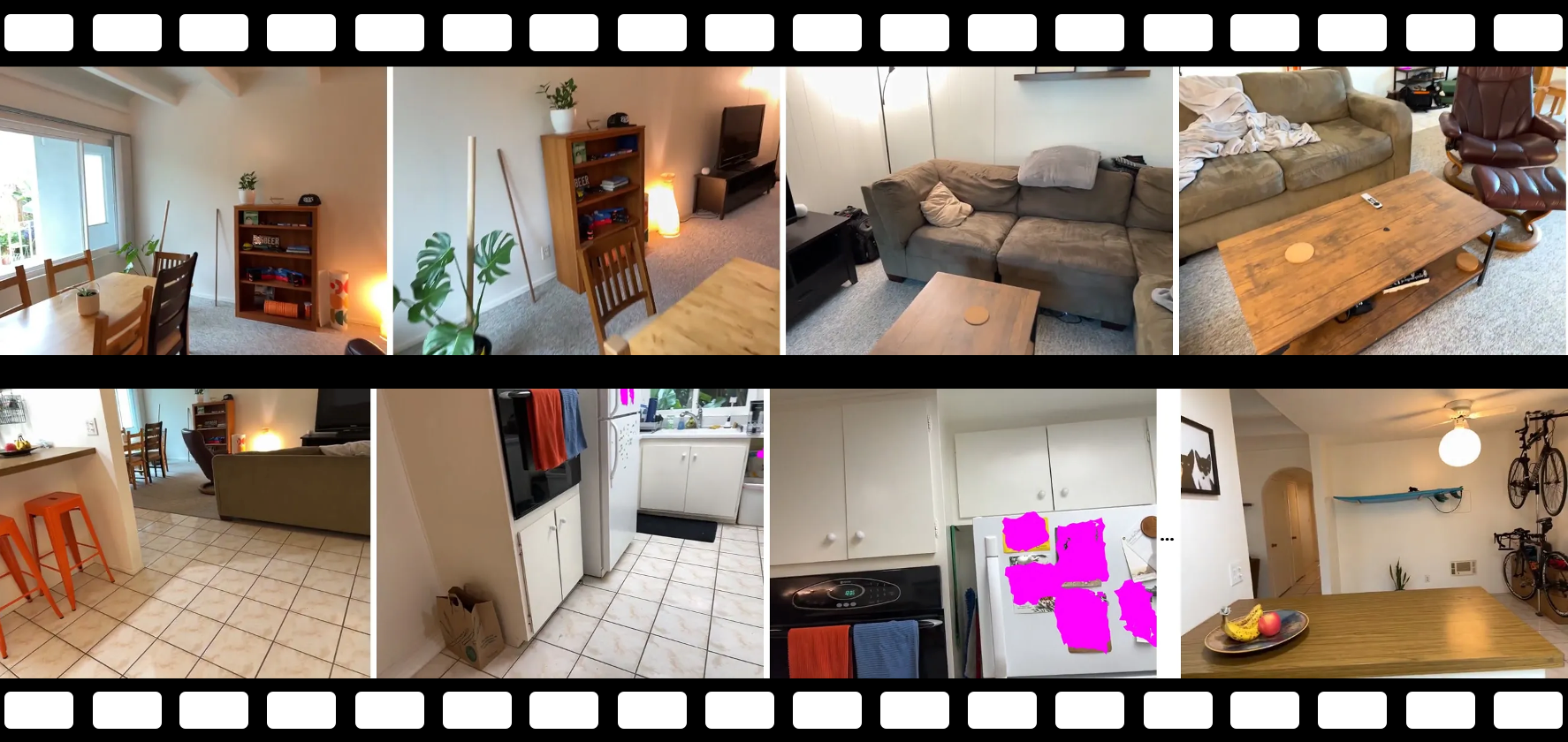} 

    \textbf{Question:} These are frames of a video. What will be the first-time appearance order of the following categories in the video: blanket, sofa, trash can, microwave? \\
    \textbf{Options:}\\
    A. blanket, sofa, trash can, microwave\\
    B. sofa, microwave, blanket, trash can\\
    C. sofa, blanket, trash can, microwave\\
    D. sofa, microwave, trash can, blanket\\
    Answer with the option's letter from the given choices directly.\\
    \textbf{GT Answer:} C

    \bigskip
    \hrule
    \bigskip

    \textbf{\method-7B:} \\
    \smallskip
    
    <think>\\
    To determine the first-time appearance order of the categories in the video, I will analyze the sequence of frames provided and identify the order in which each category (blanket, sofa, trash can, microwave) appears.\\
    \\
    1. The first frame shows a living room with a sofa, a table, and a potted plant. The sofa is visible.\\
    2. The second frame shows a dining area with a table and chairs, but no sofa, blanket, or microwave.\\
    3. The third frame shows a view of the living room again, with the sofa and a blanket on it.\\
    4. The fourth frame shows a view of the living room, with the sofa and a blanket.\\
    5. The fifth frame shows a view of the living room, with the sofa and a blanket.\\
    6. The sixth frame shows a view of the living room, with the sofa and a blanket.\\
    7. The seventh frame shows a view of the living room, with the sofa and a blanket.\\
    8. The eighth frame shows a view of the kitchen, with a trash can and a microwave.\\
    \\
    From the frames, the order of appearance is:\\
    - Sofa appears first.\\
    - Blanket appears next.\\
    - Trash can appears next.\\
    - Microwave appears last.\\
    \\
    Thus, the correct order is: sofa, blanket, trash can, microwave.\\
    </think>\\
    <answer>\\
    \boxed{C}\\
    </answer>
\end{examplebox}

\end{document}